\pdfoutput=1
\documentclass{article}
\PassOptionsToPackage{numbers,sort&compress}{natbib}
\usepackage[final,dblblindworkshop]{neurips_2026}
\workshoptitle{Interpreting Agent Behavior (IAB)}
\usepackage[T1]{fontenc}
\usepackage{graphicx}
\usepackage{booktabs}
\usepackage{array}
\usepackage{longtable}
\usepackage[expansion=false]{microtype}
\usepackage{xcolor}
\usepackage{hyperref}
\hypersetup{colorlinks=true,linkcolor=blue!55!black,citecolor=blue!55!black,urlcolor=blue!55!black,pdfauthor={Egor Pakhomov, Erik Nijkamp},pdftitle={},pdfcreator={},pdfproducer={}}
\newcolumntype{L}[1]{>{\raggedright\arraybackslash}p{#1}}

\title{Bookkeeping, Composition, or Unreachable Gold? Reading MemoryAgentBench's Conflict-Resolution Scores Against a Frozen Last-Write Resolver}
\author{%
  Egor Pakhomov \\
  Salesforce AI Research \\
  \texttt{epakhomov@salesforce.com} \\
  \And
  Erik Nijkamp \\
  Salesforce AI Research \\
  \texttt{erik.nijkamp@salesforce.com} \\
}

\begin{document}
\nocite{hu2025memoryagentbench}
\maketitle

\begin{abstract}
MemoryAgentBench's Conflict Resolution split is read as measuring ``selective forgetting''.
We execute the benchmark's own rule---the newest statement about a fact wins---as a zero-learning resolver frozen on one of the four fact lists.
Under the official metric the rule answers 80.25\% of the questions (74.5\% on the three held-out lists).
Of the rest, 67 items have a released gold that the last-write graph cannot reach but overwritten statements would (``The capital of India is New Delhi.'' superseded by ``The capital of India is Grosseto.''; gold New Delhi); such items are a third of the multi-hop questions at 262K.
Two long-context models and our pre-registered approximate re-implementation of the benchmark's BM25 agent, one retained run per item and outcomes only, score 84.7\%, 82.6\% and 41.6\% on the items the rule solves against 10.4\%, 11.9\% and 6.0\% on those 67.
The failures are a reachability split plus a small parser-scope residual; the per-item split, not the aggregate, is the unit at which a score here can be read.
\end{abstract}

\section{Introduction}\label{sec:intro}

Once a label such as ``selective forgetting'' is attached to a benchmark number, a low score reads as a missing capability and a gap between agents as a behavioral difference; whether that reading is licensed is a property of the item set and scoring rule, not of the agent \citep{raji2021everything,bowman2021fix}---a construct-validity question \citep{bean2025construct} that audits answer with trivial baselines and item-level log reading \citep{zhu2025abc,kirgis2026loganalysis}.
Partial-input and shallow-cue baselines \citep{chen2016thorough,gururangan2018annotation,poliak2018hypothesis,sugawara2018what,kaushik2018reading} and Agentless \citep{xia2024agentless} measure how far a system lacking the purported skill gets; multi-hop QA is often answerable without the composition it names \citep{min2019compositional,chen2019understanding,jiang2019avoiding}; we apply the move to an agent-memory competency. The closest work, StateMemBench \citep{fan2026statemem}, builds state-tracking probes with replay-defined gold whose traps defeat ``lazy reader'' policies---trust-most-recent first, the rule executed here; we ask the converse of an existing split: how much of it that policy already solves. Behavioral taxonomies read trajectories with an LLM coder \citep{gao2026actonomy}; we read items and outcomes.
This paper is a benchmark-score validity instrument; its behavioral evidence is outcome-only item-level runs of two long-context models and our pre-registered approximate re-implementation of the benchmark's BM25 configuration; memory systems with learned or structured stores were not run (Limitations), so for them the reinterpretation is by analogy.

MemoryAgentBench \citep{hu2025memoryagentbench} evaluates memory-equipped agents on four competencies; its Conflict Resolution split (dataset FactConsolidation) presents a numbered list of fact statements, later ones contradicting earlier ones, and asks single- and multi-hop questions about the final state.
The prompt states the rule: ``Each fact in the knowledge pool is provided with a serial number at the beginning, and the newer fact has larger serial number. You need to solve the conflicts of facts in the knowledge pool by finding the newest fact'' (v1, App.~B, Fig.~5).
The rule is executable---parse, keep the greatest position per (subject, relation) key, answer by lookup or composition---so wherever it matches gold a handcrafted zero-learning implementation suffices (bookkeeping plus our grounding and path heuristics): the item operationalizes forgetting as explicit last-write state tracking and bounds what more it demands; where it fails, an audit of \emph{why} tells what else the item demands.

\textbf{Contributions.}
(i) A frozen zero-learning last-write resolver under the official metric with a pre-declared held-out split, beside the cited published scores (\S\ref{sec:setup}--\ref{sec:published}).
(ii) A labeling of every resolver failure under a documented six-category rule (categories fixed in the study's pre-execution specification; ordering and operationalization documented at labeling time, App.~\ref{app:rule}), separating bookkeeping, misses on items whose gold the last-write graph reaches, and items whose gold it does not reach (\S\ref{sec:residual}).
(iii) Two pre-registered outcome-only experiments (two long-context models; one retrieval agent) over all 800 items, their per-item outcomes aligned to those labels (\S\ref{sec:lc}).
Neither (i) nor (ii) appears in the source paper or its repository.
Model-facing result: Table~\ref{tab:tax}'s stratification of three systems' per-item accuracy by these labels; interpretive: what a ``selective forgetting'' score can be read as measuring; we claim nothing about mechanism (one retained run per item, no trace read) or memory systems in general.

\section{Setup}\label{sec:setup}

\textbf{Corpus.}
We use the released Conflict Resolution parquet (App.~\ref{app:freeze}): 8 rows, 800 gold-answerable questions, but only \textbf{4 distinct context strings}---one fact list per tier (455 / 2,310 / 4,580 / 18,332 statements; 6K--262K after the benchmark), each appearing once with a single-hop and once with a multi-hop question set.
\textbf{Parser and resolver.}
The parser is a fixed list of 35 anchored full-line patterns over the numbered statements, mapping to 34 relations (App.~\ref{app:parser}); a statement yields a tuple only when exactly one pattern matches; per (subject, relation) key the greatest-position tuple is kept (last write).
A question is answered by literal grounding of the question entity and a graph path of at most six edges scored by fixed lexical cues and penalties; otherwise the output is \texttt{ABSTAIN}.
Resolver execution and metric scoring make no LLM API calls.

\textbf{Freeze protocol.}
Parquet rows 0 and 4 were declared the development set in the pipeline's freeze record---an internal freeze (a hashed, timed specification file; no third-party registry)---and were the only rows whose content was exported before it; they are the \emph{same} 6K fact list, so the parser was developed on \textbf{one of the four fact lists}, frozen (App.~\ref{app:freeze}), and only then run on the other three and on all 8 question sets unchanged.
On the three held-out lists the frozen parser parses 24,871 of 25,222 distinct statements (98.6\%); the 351 unparsed statements are three forms outside the pattern list (App.~\ref{app:parser}); no independent semantic audit of parses exists (App.~\ref{app:freeze}).
\textbf{Scoring.}
Headline scoring is the official repository's \texttt{substring\_exact\_match} (commit \texttt{fe1735de}; App.~\ref{app:freeze}), with \textbf{ABSTAIN counted wrong} and strict equality also reported.
Intervals are exact Clopper--Pearson \citep{clopper1934} at the answer level, conditional on the four lists; the four-list bootstrap (App.~\ref{app:sens}) is a width indicator only; clustered uncertainty is the honest scale here \citep{miller2024errorbars}.

\section{Results}\label{sec:results}

\subsection{What the frozen symbolic pipeline recovers}\label{sec:rule}

The frozen resolver answers \textbf{642/800 = 80.25\%} of questions correctly (answer-level 95\% CI 77.3--83.0; four-list interval 67.3--92.8, App.~\ref{app:sens}), with 32 ABSTAINs; held-out (six question sets), 447/600 = 74.50\% (70.8--77.9).
Per list (Table~\ref{tab:perf}, Figure~\ref{fig:acc}): single-hop 88--100\% at every length; multi-hop 95\% on the 6K list, lower on the three larger lists and lowest at 262K, whose multi-hop set contributes 64 of the 158 failures. The overwrite relation is exercised: resolving with the \emph{first} write per key instead of the last loses 526 of the 642 resolver-solved items (answered differently and scored wrong; all 266 multi-hop); 116 single-hop items are solvable under either policy, and 7 of the 67 ambiguous items have the first write as gold (App.~\ref{app:sens}).

\begin{table}[t]
\centering
\footnotesize
\setlength{\tabcolsep}{4pt}
\caption{Resolver accuracy (\%) by hop type (SH = single-hop, MH = multi-hop) and fact list ($n=100$ per list cell); pooled columns carry exact 95\% CIs (answer-level, conditional on the four lists; per-list CIs: Fig.~\ref{fig:acc}, App.~\ref{app:crosstab}); normalized (official) = strict in every cell; 6K = development list.}
\label{tab:perf}
\begin{tabular}{@{}lrrrrrr@{}}
\toprule
 & 6K (dev) & 32K & 64K & 262K & All four & Held-out (32K--262K) \\
\midrule
Single-hop & 100 & 95 & 93 & 88 & 94.0 (91.2--96.1) & 92.0 (88.3--94.8) \\
Multi-hop & 95 & 62 & 73 & 36 & 66.5 (61.6--71.1) & 57.0 (51.2--62.7) \\
ABSTAIN (SH / MH) & 0 / 0 & 2 / 7 & 5 / 4 & 4 / 10 & 11 / 21 & 11 / 21 \\
\bottomrule
\end{tabular}
\end{table}

\subsection{Position relative to published scores}\label{sec:published}

The tier-matched comparison (App.~\ref{app:published}, Table~\ref{tab:pub}; Figure~\ref{fig:acc}) is a reading aid, not a leaderboard (whole-list reading vs.\ chunked ingestion; prompting, retrieval and formatting differ---a harness-versus-model confound \citep{zhang2026harness}): at 262K the resolver's 88 / 36 exceeds every main-table system of the benchmark's arXiv v1 (at most 60 / 7; v4, the ICLR camera-ready, changes some rows: App.~\ref{app:published}), the camera-ready's GPT-5-mini (78 / 28), and GPT-4.1-mini under its always-prefer-later prompt (Policy A: 36 $\to$ 40 SH, 5 $\to$ 4 MH, $n = 100$ per cell; the camera-ready's App.~K.2, Table~19).

\subsection{The operational split: what the failures demand}\label{sec:residual}

\begin{figure}[t]
\begin{minipage}[t]{0.31\linewidth}
\vspace{0pt}
\centering
\IfFileExists{figures/fig1-half.pdf}{\includegraphics[width=\linewidth]{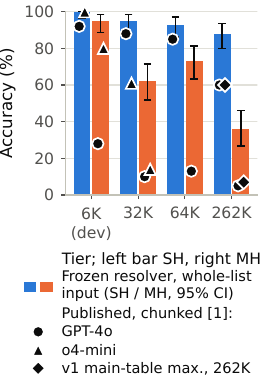}}{\IfFileExists{figures/fig1-accuracy-by-question-set.pdf}{\includegraphics[width=\linewidth]{figures/fig1-accuracy-by-question-set.pdf}}{\fbox{\parbox[c][1.6in][c]{0.94\linewidth}{\centering Figure 1 placeholder}}}}
\end{minipage}\hfill
\begin{minipage}[t]{0.68\linewidth}
\vspace{0pt}
\footnotesize
\caption{(Left.) Resolver accuracy per question set (exact 95\% CIs) and published points at their tiers (App.~\ref{app:published}; whole-list reading vs.\ chunked ingestion, not directly comparable, \S\ref{sec:published}).}
\label{fig:acc}
\centering
\setlength{\tabcolsep}{4pt}
\makeatletter\def\@captype{table}\makeatother
\caption{Operational split (\S\ref{sec:residual}; App.~\ref{app:rule}/\ref{app:examples}/\ref{app:crosstab}); cells: accuracy / share of own errors (\%), pooled over hop type (last row: accuracy on multi-hop items only, resolver-correct (266) / ambiguous (60)); Pro, Flash = Gemini 2.5; BM25 = our pre-registered approximate re-implementation of the benchmark's BM25 configuration (512-token chunks, top 10, GPT-4o-mini; \S\ref{sec:lc}, App.~\ref{app:lc}).}
\label{tab:tax}
\begin{tabular}{@{}lrrr@{}}
\toprule
Item class ($n$; share of 158 failures) & Pro & Flash & BM25 \\
\midrule
Resolver-correct (642) & 84.7 / 49.0 & 82.6 / 54.1 & 41.6 / 72.8 \\
Last-write-reachable miss (78; 49.4\%) & 52.6 / 18.5 & 59.0 / 15.5 & 12.8 / 13.2 \\
Ambiguous gold, oper.\ (67; 42.4\%) & 10.4 / 30.0 & 11.9 / 28.5 & 6.0 / 12.2 \\
Parser scope (13; 8.2\%) & 61.5 / 2.5 & 69.2 / 1.9 & 30.8 / 1.7 \\
All (800) & 75.0 / 100 & 74.1 / 100 & 35.6 / 100 \\
\midrule
Multi-hop only (266 / 60) & 74.4 / 8.3 & 66.2 / 11.7 & 11.7 / 0.0 \\
\bottomrule
\end{tabular}
\end{minipage}
\end{figure}

One row (App.~\ref{app:examples}, B5): multi-hop 6K \#67 asks for the capital of the country that gave rise to Rog\'erio Ceni's sport; statement 158 reads ``The capital of India is New Delhi.'', statement 341 reads ``The capital of India is Grosseto.'', and the released gold is New Delhi---the earlier, overwritten value.

\textbf{Label production.}
The 158 failures were labeled by a classification script over a row-level evidence workbench: 13 rows were designated as parser-scope failures in an LLM-assisted batch review under the authors' direction (a required query edge occurs only in an unparsed statement; App.~\ref{app:rule}); the script labeled every other row from the workbench's last-write reachability flag---reachable $\to$ \emph{last-write-reachable miss}; not reachable $\to$ \emph{ambiguous gold}.
Its override lists for the other three categories were left empty; category names were pre-specified; ordering, the reachability operationalization and the 13 designations were fixed during labeling (App.~\ref{app:rule}); 158/158 labeled (Table~\ref{tab:tax}).

\emph{Every failure's gold occurs in the context, so the empty category follows from the rule:} ``genuine beyond last write'' can hold only an item whose gold is absent from the context (App.~\ref{app:rule}).

\emph{Half the residue: last-write-reachable misses.}
78 failures (49.4\%; 67 multi-hop, 11 single-hop, 22 ABSTAINs) have a gold reachable from a question mention by \emph{some} last-write path of at most six edges---the flag does not test that the path answers the question---and the resolver did not return it; we call them last-write-reachable misses (mostly, we expect, grounding or path-selection errors of our resolver, which a stronger grounder could reduce; the flag alone does not establish that).
8 of the 78 fail a cue-consistency screen (App.~\ref{app:rule}), one of them (multi-hop 262K \#77) clearly.

\emph{Two-fifths of the residue is operationally ambiguous gold.}
67 failures (42.4\%; none ABSTAIN) have a released gold that occurs in the context but is not reachable from a question mention in the frozen last-write graph; for all 67 it becomes reachable within six edges once overwritten tuples are re-admitted (base rate 25\% for other questions' golds, App.~\ref{app:rule})---in 50 the gold-valued tuple itself was overwritten, in 17 only an intermediate one; a row-by-row reading confirms all 67 (App.~\ref{app:rule}).
The label is operational (relative to the frozen parser's graph), not a claim that the released gold is wrong; 3 of the 67 also have the gold inside an unparsed statement (App.~\ref{app:examples}; a fuller parser could relabel two).
App.~\ref{app:examples} collects the provenance of these items: MQuAKE and its known conflict issues \citep{zhong2023mquake,mquake_remastered,mquake2024readme}, ripple effects of edits \citep{cohen2024ripple}, and in-context knowledge conflicts \citep{longpre2021entity,xu2024knowledgeconflicts}.
Any procedure answering from the exact last-write graph is scored incorrect on them (66 of 67 under the substring surface, App.~\ref{app:rule}); 60 of 67 are multi-hop, concentrated at 262K (App.~\ref{app:crosstab}). The remaining 13 failures (8.2\%) need an edge found only in an unparsed statement (App.~\ref{app:examples}).

\subsection{Three systems, item by item}\label{sec:lc}

We ran Gemini 2.5 Pro on all 800 questions under the benchmark's long-context prompt and assembly (deviations: App.~\ref{app:lc}), one retained scored output per item (pre-registered; outcome stratification, not mechanism evidence); it scores 75.0\% (single-hop 90.0\%, multi-hop 60.0\%; held-out 68.5\%; CIs in App.~\ref{app:lc}).
Per class (Table~\ref{tab:tax}): 84.7\% vs.\ 10.4\% on resolver-solved vs.\ ambiguous items (8.4\% of questions, 30\% of its 200 errors).
Its largest error block is elsewhere: 98 misses on items the rule solves, 61 at 262K (Flash: 112 and 40)---consistent with, but not a test of, length-dependent degradation \citep{liu2024lost,hsieh2024ruler}.
Gemini 2.5 Flash reproduces the class ordering, not the per-tier values (Table~\ref{tab:tax}; $\kappa = 0.53$ with Pro).
Our pre-registered approximate BM25 re-implementation (App.~\ref{app:lc}) scores 35.6\% (62.8\% single-hop, 8.5\% multi-hop); its top-10 chunk set matches the closest official-code variant's on 134 of 800 items, so 56 / 4 at 262K vs.\ the published 56 / 3 is only a coarse fidelity check.
Per class it follows the same ordering (Table~\ref{tab:tax}): 6.0\% (4/67) on the operationally ambiguous items (12\% of its errors), 73\% of its errors on items the rule solves.
For the retrieval agent, 496 of 800 items had every resolver-used statement in the top-10 chunks yet only 262 were answered correctly, and a gold-valued statement was retrieved for 336 of its 515 errors: the resolver's statements were often retrieved, but with outcomes only the post-retrieval cause cannot be localized.
Class membership is confounded with hop type and with length (60 of the 67 ambiguous items are multi-hop, 34 of them at 262K); within multi-hop items, pooled over lengths, resolver-solved vs.\ ambiguous accuracy is 74.4\% vs.\ 8.3\% (Pro, 198/266 vs.\ 5/60), 66.2\% vs.\ 11.7\% (Flash, 176/266 vs.\ 7/60) and 11.7\% vs.\ 0\% (agent, 31/266 vs.\ 0/60): the pooled gap persists for the long-context models but shrinks to 12 points for the agent; at multi-hop 262K alone it is 4/36 vs.\ 1/34 for Pro (App.~\ref{app:lc}).
Three runs of the retrieval agent under the identical protocol put its accuracy on the operationally ambiguous items at 6.0--9.0\% and on resolver-solved items at 40.3--41.7\%, with per-item agreement between runs $\kappa$ = 0.90 (35--37 of 800 items change outcome); the Gemini scores are single runs (App.~\ref{app:lc}).

\section{Discussion}\label{sec:discussion}

On this split, \emph{single-hop}: 94.0\% of items (92.0\% held-out) are solved by one-edge greatest-position lookup, and the two long-context models solve 90\% and 92\%.
\emph{Multi-hop}: composition over a growing graph plus about a sixth of the multi-hop half (a third at 262K) whose gold the rule cannot reach, on which accuracy falls to 8.3\%, 11.7\% and 0\% (\S\ref{sec:lc}); under the documented rule no multi-hop miss was attributed to semantics beyond last write (\S\ref{sec:residual})---true of this split, not of conflict resolution in general, for which last-write-defeating probes exist \citep{fan2026statemem}; read multi-hop scores against the ambiguous fraction per tier (App.~\ref{app:crosstab}).

\textbf{Limitations.}
(1) \emph{Four fact lists, no memory store}: intervals are conditional on the four lists (App.~\ref{app:sens}); no memory system with a learned or structured store was analyzed at the item level.
(2) \emph{Label production}: 145 of the 158 labels follow from the reachability flag; the 13 parser-scope rows were designated in an LLM-assisted batch review (GPT-5.6) and are the sole discretionary labels: a re-labeling blind to the frozen labels (flag shown) agreed on 147/158 ($\kappa$ 0.88 \citep{cohen1960kappa}) but on 6 of the 13, so agreement on the 145 measures rule-following; one without the explicit flag (evidence capped and path-seeded; last-write tags visible) agreed on 141/158 ($\kappa$ 0.82) and recovered 12 of 13; no per-row record of the review is available; the mechanical criterion is post hoc; the frozen labels are unchanged.
(3) \emph{One split, one rule, one parser}: the 35-pattern parser is a lower bound on a symbolic procedure; its heuristics are ours (App.~\ref{app:freeze}).
(4) \emph{Within-corpus held-out}: developed on the 6K list; the held-out lists share its templates.
\textbf{Agenda.}
Align agents' traces, not just outcomes, with the labels; re-adjudicate the 67 items with the maintainers; test other splits. LLM use: App.~\ref{app:llm}.

\begin{ack}
This work was funded by Salesforce.
\end{ack}

\bibliographystyle{unsrtnat}
\bibliography{main}

\appendix

\section{Taxonomy rule, label production, and reachability check}\label{app:rule}

\textbf{Decision rule (six categories fixed before execution in the study's pre-execution specification---internally the ``gate specification'', an excerpt of which is available from the authors on request; the ordering and the reachability operationalization were documented in the analysis log (the ``gate report'') by the LLM-based pipeline that operated the analysis under the authors' direction (the ``operating agent''; App.~\ref{app:llm}) when the labels were produced).}
\begin{quote}
\small
required query edge present only in a frozen-parser unparsed statement $\to$ parse failure;
released gold inconsistent with or unreachable under greatest-position overwrite despite context occurrence $\to$ ambiguous gold;
gold reachable in the frozen last-write graph but query grounding/path selection chooses another output $\to$ multi-hop composition failure;
demonstrable semantics not expressible by greatest-position overwrite $\to$ genuine beyond-last-write;
scoring-surface-only difference $\to$ normalization artifact;
no isolated cause $\to$ unclassifiable.
\end{quote}
The rule's names are quoted as frozen; this paper writes \emph{parse failure} as \emph{parser scope} and \emph{multi-hop composition failure} as \emph{last-write-reachable miss} (the label strings in the audit file stay \texttt{parse\_failure} and \texttt{multi\_hop\_composition\_failure}; the second name predates the finding that the class holds 11 single-hop rows), and uses no other aliases.

\textbf{Workbench.}
For each of the 158 headline failures a row-level evidence workbench was built by script before any label existed; the workbench file (available on request) is a deterministic regeneration of the one shown to the labeler in batches.
Each row carries the question, the released gold strings, the resolver's prediction and trace reason, the predicted start entity, the predicted relation path and the statements it used, the entity mentions and relation/terminal cues found in the question, whether the gold is reachable from some question mention in the frozen last-write graph, every gold-reaching path with its statement indices, the parsed context statements containing the gold string, the unparsed context statements containing the gold string, and every context line containing the gold string.

\textbf{Label production.}
The labels were produced by the classification script over this workbench.
In an LLM-assisted review (GPT-5.6) directed by the authors, the workbench was read in batches and 13 rows (residual ids---row numbers in the 158-row audit file---13, 17, 30, 48, 56, 67, 108, 138, 141, 142, 153, 157, 158) were designated as parse failures, on the ground that a required query edge occurs only in an unparsed statement; these ids are an explicit list in the script.
A post hoc mechanical proxy---whether an unparsed statement's whole subject phrase occurs in the question text (whole-word match after the official normalization, tolerating a possessive s, e.g.\ ``Attorney General's'')---recovers exactly the 13 designated rows and no other failure.
The proxy was formulated after the designations, to test them; it is not part of the documented rule, and it does not test whether the matched statement supplies a required query relation or whether the subject lacks other parsed outgoing edges (two designated subjects have parsed edges of other relations): the required-edge reading of these rows comes from the designation review.
Every other row was labeled by the script from the workbench's last-write reachability flag: gold reachable in the frozen last-write graph $\to$ last-write-reachable miss (78 rows; the script's label string is \texttt{multi\_hop\_composition\_failure}, kept in the audit file; earlier drafts called the class grounding/path-selection failure; the category holds 67 multi-hop and 11 single-hop rows, 22 of them ABSTAINs); not reachable $\to$ ambiguous gold (67 rows). The flag tests reachability from \emph{some} question mention by \emph{any} relation sequence of at most six edges, without regard to the chain the question asks for: all 78 last-write-reachable-miss rows are reachable from a question mention, 56 of them from the resolver's chosen start entity (the other 22 are ABSTAINs with no start entity).
A screen requiring some gold-reaching last-write path to contain every relation the resolver cued from the question flags 8 of the 78; in one of them (multi-hop 262K \#77) the resolver's own path covers every cued relation and reaches the last-write answer while the gold sits on a relation sequence the question does not cue (author $\to$ died\_city), so under a chain-consistent test that row is operationally ambiguous; the other seven are ABSTAINs or paths that themselves miss a cue.
The two models' accuracy on the 8 screen-flagged rows is 5/8 (Pro) and 4/8 (Flash) versus 36/70 and 42/70 on the other 70 last-write-reachable-miss rows.
The frozen labels are reported unchanged.
The 11 single-hop rows in this category are resolver defects: five ABSTAIN with an empty cue set (no entry of the frozen cue table matches the question form ``Which language was X written in?''), and six prefer a 3--4-edge path over the 1-edge gold path because a cued relation is credited each time it appears on a path.
The script also carries explicit override lists for the categories \emph{genuine conflict resolution beyond last write}, \emph{normalization artifact} and \emph{unclassifiable}; all three were left empty.
The script writes one uniform audit note per label:
last-write-reachable miss---``gold is reachable in the frozen last-write graph, but query grounding/path selection chose another output'';
ambiguous gold---``released gold is not reachable from the question entity in the frozen last-write graph despite appearing in the context'';
parse failure---``required query edge appears only in an unparsed statement''.
158/158 rows received a label; 0 were unclassifiable.
The pre-execution specification called for every failure to be classified by the operating agent; what was executed was a script over the reachability flag plus 13 agent designations---a deviation we disclose.

\textbf{Blind re-adjudication.}
After the labels were frozen, a second language model, of a different family from the one used in the batch review (Claude Opus 4.8), received the same eight workbench batches (which contain no label) and the same six-category rule in the same order, and returned a label and a one-sentence reason for every row.
Agreement with the frozen labels: 147/158 (93.0\%), Cohen's $\kappa = 0.875$.
By frozen label: last-write-reachable miss 76/78, ambiguous gold 65/67, parse failure 6/13.
Confusion: 2 last-write-reachable-miss rows $\to$ ambiguous gold (the second labeler required the gold-reaching path to match the question's relation chain); 1 ambiguous-gold row $\to$ last-write-reachable miss; 1 ambiguous-gold row $\to$ parse failure (the row whose gold occurs in 31 unparsed ``original language'' statements); 7 parse-failure rows $\to$ ambiguous gold: all 7 satisfy the mechanical criterion above (their first hop---the head coach of a club, the President of a country, the pope, a state attorney general---exists only in unparsed statements); the second labeler applied the reachability test before the parse-scope test, and in several of its reasons it notes the missing head-coach or office edge and still chooses ambiguous gold (e.g., ``gold\_reachable\_lww False and no head coach edge grounds to Australia's sport, so gold unreachable under overwrite''), so these are disagreements about the order in which the rule's tests are applied.
The per-row labels and reasons are recorded in the audit file, available on request.
No frozen label was changed.
Because 145 of the 158 reference labels are a deterministic function of the reachability flag, which the second labeler saw, agreement on those rows measures rule-following; the informative part is the 13 parser-scope rows.
\textbf{Re-adjudication B.}
Because re-adjudication A showed the labeler the reachability flag that determines 145 of the 158 reference labels, a second blind re-labeling withheld the explicit flag. A third model family (Gemini 3.1 Pro (preview)) saw, per row, the question, gold, resolver output and path, and a set of candidate context statements, and applied the same six-category rule. The candidate builder selected statements mentioning a question, path or gold entity and, where present, entities on the computed gold-reaching paths, so evidence selection was itself reachability-conditioned; it targeted 60 candidates per row while retaining every unparsed candidate (one row therefore retained 121); each candidate was tagged parsed (with subject, relation, object and whether it is the last write for its key) or unparsed.
Agreement with the frozen labels: 141/158 (89.2\%), Cohen's $\kappa = 0.82$.
It labeled 12 of the 13 designated parser-scope rows parse failure and one further row (a last-write-reachable miss under the frozen labels) parse failure; 8 rows it called unclassifiable: 7 on rows whose evidence hit the cap (2 of them say so), 1 (\#77) on an untruncated row whose block showed the gold-reaching died\_city path but not the spouse $\to$ citizenship edges the labeler reported absent, which the workbench's triggers omit; 6 ambiguous-gold rows it called last-write-reachable misses, reasoning about paths whose last-write status it misjudged; one last-write-reachable-miss row it called ambiguous gold and the remaining designated parser-scope row a last-write-reachable miss (17 disagreements in all); the two re-labelings agree with each other on 134/158 ($\kappa = 0.74$).
Its disagreements concentrate on rows whose evidence hit the 60-statement cap: 52 of 158 rows hit the cap; 7 of the 8 unclassifiable rows and 14 of the 17 disagreements are truncated rows (disagreement 26.9\% on truncated vs 2.8\% on untruncated rows).
The per-statement last-write tags remained visible, so a diligent labeler could reconstruct aggregate reachability; only the precomputed flag was withheld.
Frozen labels unchanged.

\textbf{Construction note.}
Every one of the 158 rows has the gold string in at least one context line.
Given the rule's ordering, an item whose gold occurs in the context is assigned to parse failure, ambiguous gold or multi-hop composition failure before ``genuine beyond-last-write'' is considered; that category can therefore be reached only by an item whose gold is absent from the context, and under the documented rule it is empty (\S\ref{sec:residual}).

\textbf{Additional per-item reachability check (not part of the labeling).}
For each ambiguous-gold row, a breadth-first search over subject $\to$ object edges of \emph{all} parsed tuples of the context (overwritten tuples re-admitted), depth $\leq 6$, from the resolver's start entity, reaches a gold string under the official normalization in 67/67 rows, and in 0/67 rows when only last-write tuples are used (reproducing the workbench flag).
In 50 of the 67 some re-admitted path of at most six edges ends in a gold-valued tuple that a later statement overwrote (existence over paths); in the other 17 only intermediate tuples on such paths were overwritten and the gold-valued tuple at the end of the path is itself a last write; on the shortest re-admitted path the split depends on tie-breaking and the matching surface, so we report the existence version only.
Reachability here tests exact equality of the normalized node with a normalized gold, as the workbench does; under the scoring rule's substring test one row (multi-hop 262K, gold ``Vienna'') is reachable at depth one through the unrelated node ``Academy of Fine Arts Vienna'', so the count is 66/67 under that looser test.
Base rate: from the same start entities, the golds of the other 99 questions in the same set are reachable within six edges in 5.6\% of pairs under last-write and 25.4\% once overwritten tuples are re-admitted (6,633 pairs), and sit at the end of a superseded edge on some such path in 12.7\%; the corresponding rates for the 67 rows' own golds are 1.5\% (1/67), 100\% (67/67) and 74.6\% (50/67) (substring surface throughout).

\textbf{Row-by-row check of the 67 rows.} A workbench (available on request) lists, for each of the 67 rows, the question, the released gold, the resolver's answer and the statements it used, every parsed statement whose object is the gold together with the last-write statement on the same key, and the shortest gold-reaching path once overwritten statements are re-admitted. Two models read it independently under the authors' direction (Claude Opus 4.8 with file access; GPT-5.6 with the workbench inlined) and judged, row by row, whether the released gold is a value that a later statement on the same key superseded, or is reachable only through such a statement: 67 of 67 confirmed by both, no disagreement (verdict files available on request). This checks the raw statements, not the parser's tuples, so it does not replace the parse audit that App.~\ref{app:freeze} says is missing.
The re-admitted graph reaches a median of 47 nodes from a start entity (87 at 262K) against 8 under last-write.
Per question set: multi-hop 262K 34, 32K 13, 64K 9, 6K 4; single-hop 262K 3, 64K 2, 32K 2.
All 158 failures have at least one context line containing the gold string.
Three ambiguous-gold rows also have the gold string inside an unparsed statement (residual ids 73, 86, 132; App.~\ref{app:examples}).

\section{Examples per category}\label{app:examples}

Each example gives question $\to$ gold $\to$ resolver output, with the question set and question index.

\textbf{Last-write-reachable miss.}
(B1) ``From which country does the musical genre Michael Mantler is associated with come from?'' $\to$ Canada $\to$ post-punk (multi-hop 6K, \#38; the path stopped one edge short---gold is reachable via \texttt{music\_type} $\to$ \texttt{created\_country}, statements 298 and 267).
(B2) ``What is the occupation of the chairperson of the Palestine Liberation Organization?'' $\to$ basketball player $\to$ Regina Ip (multi-hop 32K, \#4).
(B3) ``Who holds the title of chief public representative in the country where Jason Motte is a citizen?'' $\to$ Roch Marc Christian Kabor\'e $\to$ Ireland (multi-hop 32K, \#7).

\textbf{Ambiguous gold (operational).}
(B4) ``Which city serves as the capital of the country where the sport that Mike D'Antoni specialized in hails from?'' $\to$ London $\to$ Oderzo (multi-hop 6K, \#44).
(B5) ``Which city serves as the capital of the country that gave rise to the sport played by Rog\'erio Ceni?'' $\to$ New Delhi $\to$ Grosseto (multi-hop 6K, \#67; context statements 158 ``The capital of India is New Delhi.'' and 341 ``The capital of India is Grosseto.'').
(B6) ``What is the name of the capital city of the country where the founder of the manufacturer of Ford F-Series hailed from?'' $\to$ New Delhi $\to$ Grosseto (multi-hop 6K, \#86).
The three rows whose gold string also occurs in an unparsed statement: multi-hop 262K \#4 (gold ``English''; 31 unparsed occurrences, ``The original language of \ldots\ is English.'' forms), multi-hop 262K \#20 (gold ``Sergio Mattarella''; unparsed ``The President of Italy is Sergio Mattarella.''), multi-hop 262K \#96 (gold ``Australia''; inside the subject phrase of two unparsed head-coach statements, ``The head coach of Football Federation Australia is \ldots'', where it is not a value).

\textbf{Provenance of the fact lists.}
The benchmark's construction statement (v1 App.~A.4; identical in the v4 camera-ready's App.~B) says: ``For each edit pair, the sentence representing outdated information (the distractor) is given a smaller number, while the sentence representing more recent information (the one containing the answer) is given a larger number.''
The 50 rows in which a later statement on the same key supersedes the gold-valued statement are the concrete content of ``operationally ambiguous'' relative to this construction; neither v1 nor v4 names the MQuAKE subset used.
The MQuAKE repository's README documents a knowledge-conflict issue in MQuAKE-CF-3k under multi-edit use and points to the corrected subset \citep{mquake2024readme}; a later audit found a large share of MQuAKE questions and labels corrupted and its multi-hop editing evaluations unreliable without repairs \citep{mquake_remastered}; and conflicting evidence---between a context and parametric memory \citep{longpre2021entity}, or between statements inside one context \citep{xu2024knowledgeconflicts}---is a recognized failure surface for language models in its own right.
Whatever their origin, the 67 operationally ambiguous items are ones on which a procedure that reproduces the frozen last-write graph is scored incorrect under exact node match (66 of 67 under the substring surface, App.~\ref{app:rule}).

\textbf{Parser scope.}
(B7) ``What is the country of origin of the sport to which the head coach of Western Sydney Wanderers FC belongs?'' $\to$ Australia $\to$ ABSTAIN (multi-hop 32K, \#23; ``head coach'' form unparsed).
(B8) ``What is the name of the partner of the President of Azerbaijan?'' $\to$ Mahidol Adulyadej $\to$ ABSTAIN (multi-hop 32K, \#29; titled-office form unparsed).
(B9) ``What is the name of the capital of the country where the sport coached by the head coach of the Cleveland Indians was invented?'' $\to$ Amposta $\to$ Grosseto (multi-hop 32K, \#70).

\section{Parser and resolver}\label{app:parser}

\textbf{Statement segmentation.}
Each line of the context is tested against the full-line regular expression \verb|\s*(\d+)\.\s+(.+?)\s*|; a matching line contributes one statement with the captured integer as its numbered position.
The unnumbered heading is excluded.

\textbf{Patterns.}
Table~\ref{tab:patterns} lists the 35 frozen patterns in their order of application, with the relation key each yields.
Every pattern is applied as a full-line match against the statement text; a statement yields a tuple (subject $=$ group \texttt{s}, relation, object $=$ group \texttt{o}, position) only when \emph{exactly one} pattern matches, and is otherwise recorded as unparsed.
The order is deliberate: more syntactically specific patterns precede generic ones.
Two surface forms (\texttt{head\_government} and the ``Prime Minister'' form) share the relation key \texttt{head\_government}, so the 35 patterns map to 34 relation keys.

\begin{table}[htbp]
\centering
\footnotesize
\setlength{\tabcolsep}{4pt}
\caption{The 35 frozen full-line patterns (Python \texttt{re} syntax), listed in order of application, with the relation key each yields; spellings such as ``univeristy'' and ``origianl'' reproduce the benchmark's statement templates.}
\label{tab:patterns}
\begin{tabular}{@{}ll@{}}
\toprule
Relation key & Pattern \\
\midrule
\verb|chairperson| & \verb|The chairperson of (?P<s>.+) is (?P<o>.+)\.| \\
\verb|director| & \verb|The director of (?P<s>.+) is (?P<o>.+)\.| \\
\verb|headquarters_city| & \verb|The headquarters of (?P<s>.+) is located in the city of (?P<o>.+)\.| \\
\verb|author| & \verb|The author of (?P<s>.+) is (?P<o>.+)\.| \\
\verb|ceo| & \verb|The chief executive officer of (?P<s>.+) is (?P<o>.+)\.| \\
\verb|educated_at| & \verb|The univeristy where (?P<s>.+) was educated is (?P<o>.+)\.| \\
\verb|capital| & \verb|The capital of (?P<s>.+) is (?P<o>.+)\.| \\
\verb|head_state| & \verb|The name of the current head of state in (?P<s>.+) is (?P<o>.+)\.| \\
\verb|head_government| & \verb|The name of the current head of the (?P<s>.+) government is (?P<o>.+)\.| \\
\verb|head_government| & \verb|The Prime Minister of (?P<s>.+) is (?P<o>.+)\.| \\
\verb|music_type| & \verb|The type of music that (?P<s>.+) plays is (?P<o>.+)\.| \\
\verb|official_language| & \verb|The official language of (?P<s>.+) is (?P<o>.+)\.| \\
\verb|produced_by| & \verb|The company that produced (?P<s>.+) is (?P<o>.+)\.| \\
\verb|original_broadcaster| & \verb|The origianl broadcaster of (?P<s>.+) is (?P<o>.+)\.| \\
\verb|born_city| & \verb|(?P<s>.+) was born in the city of (?P<o>.+)\.| \\
\verb|died_city| & \verb|(?P<s>.+) died in the city of (?P<o>.+)\.| \\
\verb|plays_position| & \verb|(?P<s>.+) plays the position of (?P<o>.+)\.| \\
\verb|located_continent| & \verb|(?P<s>.+) is located in the continent of (?P<o>.+)\.| \\
\verb|worked_city| & \verb|(?P<s>.+) worked in the city of (?P<o>.+)\.| \\
\verb|spouse| & \verb|(?P<s>.+) is married to (?P<o>.+)\.| \\
\verb|founded_by| & \verb|(?P<s>.+) was founded by (?P<o>.+)\.| \\
\verb|associated_sport| & \verb|(?P<s>.+) is associated with the sport of (?P<o>.+)\.| \\
\verb|citizen_country| & \verb|(?P<s>.+) is a citizen of (?P<o>.+)\.| \\
\verb|performed_by| & \verb|(?P<s>.+) was performed by (?P<o>.+)\.| \\
\verb|founded_city| & \verb|(?P<s>.+) was founded in the city of (?P<o>.+)\.| \\
\verb|created_country| & \verb|(?P<s>.+) was created in the country of (?P<o>.+)\.| \\
\verb|child| & \verb|(?P<s>.+)'s child is (?P<o>.+)\.| \\
\verb|created_by| & \verb|(?P<s>.+) was created by (?P<o>.+)\.| \\
\verb|religion| & \verb|(?P<s>.+) is affiliated with the religion of (?P<o>.+)\.| \\
\verb|famous_for| & \verb|(?P<s>.+) is famous for (?P<o>.+)\.| \\
\verb|employed_by| & \verb|(?P<s>.+) is employed by (?P<o>.+)\.| \\
\verb|speaks_language| & \verb|(?P<s>.+) speaks the language of (?P<o>.+)\.| \\
\verb|work_field| & \verb|(?P<s>.+) works in the field of (?P<o>.+)\.| \\
\verb|developed_by| & \verb|(?P<s>.+) was developed by (?P<o>.+)\.| \\
\verb|written_language| & \verb|(?P<s>.+) was written in the language of (?P<o>.+)\.| \\
\bottomrule
\end{tabular}
\end{table}

\textbf{Last-write graph.}
For each (subject, relation) key, the tuple with the greatest numbered position is retained; all earlier tuples for that key are discarded.
Subjects and objects are used as written (after stripping surrounding whitespace); no entity linking is performed.

\textbf{Entity grounding.}
Every subject and object of the last-write graph is a node.
A node is a mention of the question if its normalized surface (Unicode NFKC, case-folded, curly quotes unified, whitespace collapsed) occurs as a substring of the normalized question; when mentions overlap, a longer literal dominates any literal whose span lies within it.

\textbf{Path search and scoring.}
From each mention, all cycle-free paths of 1 to 6 edges in the last-write graph are enumerated (outgoing edges ordered by relation, then object).
Each path receives a deterministic score:
$\min(\ell,60)/100$, where $\ell$ is the mention length in characters;
plus, for each relation on the path, the weight of the strongest fixed lexical cue for that relation found in the question (a fixed table of regular expressions per relation, weights 1.0--5.0);
plus a fixed terminal preference of 12.0 when the question matches one of a fixed list of question constructions associated with the path's last relation;
minus $0.55$ per edge;
minus $0.45$ per relation on the path with no cue in the question;
plus $0.35$ per distinct cued relation on the path;
minus $6.0$ if the terminal value's normalized surface already occurs in the question.
The highest-scoring path's terminal value is the answer; ties are broken in favour of the later mention position in the question (larger start index), then by terminal value, then by relation sequence.
The cue and preference tables are fixed in the frozen source and were not changed after the freeze.

\textbf{ABSTAIN conditions.}
The resolver outputs the literal \texttt{ABSTAIN} when no node is mentioned in the question or no path leaves any mentioned node, or when the best path's score is $\leq 0$, or when the question triggers no relation cue and no terminal preference.
ABSTAIN is scored as wrong in every headline number.

\textbf{Unparsed-statement census.}
Over the three distinct held-out fact lists (32K, 64K, 262K: 2,310 + 4,580 + 18,332 = 25,222 statements), the frozen parser leaves 351 statements unparsed (29 + 60 + 262): 166 of the form ``The head coach of X is Y'', 69 of the form ``The original language of X is Y'', and 116 titled-office statements of the form ``The $\langle$office$\rangle$ is $\langle$person$\rangle$'' (e.g., ``The President of Azerbaijan is Ilham Aliyev.'').
Each held-out list occurs in two parquet rows, so the per-row totals are 702 unparsed of 50,444 (49,742 parsed, 98.6\%).
The development list (455 statements) is parsed in full.

\section{Sensitivities}\label{app:sens}

\begin{table}[htbp]
\centering
\footnotesize
\caption{Sensitivity of the headline accuracy. All entries are normalized (official) scoring; strict scoring gives identical counts in every row.}
\label{tab:sens}
\begin{tabular}{@{}L{2.9in}L{2.3in}@{}}
\toprule
Variant & Accuracy (\%) \\
\midrule
All 800 answers, ABSTAIN wrong (headline) & 642/800 = 80.25 \\
ABSTAIN excluded (768 answers) & 642/768 = 83.59 \\
Held-out six question sets only (three fact lists) & 447/600 = 74.50 \\
Leave one question set out (8 sets = 4 lists $\times$ 2 hop types), omitted set 0--7 & 78.14 / 82.86 / 81.29 / 86.57 / 77.43 / 78.14 / 78.43 / 79.14 \\
\quad range & 77.43--86.57 (width 9.14 pp) \\
Leave one fact list out (both of its question sets), dropped 6K / 32K / 64K / 262K & 74.50 / 80.83 / 79.33 / 86.33 \\
Fact-list-clustered percentile bootstrap, 4 clusters of 200 answers, 10,000 replicates, seed 20260901 & 80.25 [67.25, 92.75]; half-width 12.75 pp \\
\bottomrule
\end{tabular}
\end{table}

\paragraph{Overwrite policy.} Re-resolving every question with the frozen parser and heuristics but keeping the \emph{first} write per (subject, relation) key instead of the last (analysis script available on request; the last-write rerun reproduces 642/800 exactly): 526 of the 642 resolver-solved items are lost under first-write---answered differently and scored wrong (260 single-hop, all 266 multi-hop; one further single-hop item is answered differently but both answers contain the gold), 116 single-hop items are solved under both policies (115 because every key on their path has a single write; one because both policies' answers contain the gold), 7 of the 67 operationally ambiguous items are solved only under first-write (their gold is the earlier value), and the remaining 151 failures are solved by neither. The split therefore does exercise the overwrite relation on 82\% of the items the rule solves; what it does not require is anything beyond taking the newest write.

Parquet rows 0--7 are, in order, multi-hop 6K, 32K, 64K, 262K and single-hop 6K, 32K, 64K, 262K.
With only 4 clusters the bootstrap resampling distribution has 256 equiprobable outcomes; the interval is reported as a width indicator only and supports no precision claim.
The frozen gate record's clustered bootstrap used the 8 question sets as clusters and gave [65.1, 93.0]; because the 8 sets are 4 fact lists $\times$ 2 question sets, the paper reports the 4-list version, [67.25, 92.75].

\section{Accuracy grid with intervals; residual splits}\label{app:crosstab}
Table~\ref{tab:perfci} repeats the accuracy grid of Table~\ref{tab:perf} with exact 95\% intervals; Table~\ref{tab:crosstab} splits the 158 failures by question set and class, and Table~\ref{tab:taxsplit} cross-tabulates them by hop and output type.

\begin{table}[htbp]
\centering
\footnotesize
\caption{Residual taxonomy of the 158 failures by hop type and by output type (counts from the row-level audit file); Model correct = Gemini 2.5 Pro (\S\ref{sec:lc}).}
\label{tab:taxsplit}
\begin{tabular}{@{}lrrrr@{}}
\toprule
Label & $n$ (share) & Multi-hop / single-hop & ABSTAIN / committed & Model correct \\
\midrule
Last-write-reachable miss & 78 (49.4\%) & 67 / 11 & 22 / 56 & 41 \\
Ambiguous gold (operational) & 67 (42.4\%) & 60 / 7 & 0 / 67 & 7 \\
Parser scope & 13 (8.2\%) & 7 / 6 & 10 / 3 & 8 \\
Genuine beyond last write & 0 & -- & -- & -- \\
Normalization artifact & 0 & -- & -- & -- \\
Unclassifiable & 0 & -- & -- & -- \\
\bottomrule
\end{tabular}
\end{table}

\begin{table}[htbp]
\centering
\footnotesize
\setlength{\tabcolsep}{4pt}
\caption{Per-list accuracy (\%) with exact Clopper--Pearson 95\% CIs ($n=100$ per hop cell, 200 per both-hops cell); the pooled and held-out columns with their CIs are in Table~\ref{tab:perf}.}
\label{tab:perfci}
\begin{tabular}{@{}lrrrr@{}}
\toprule
 & 6K (dev) & 32K & 64K & 262K \\
\midrule
Single-hop & 100 (96.4--100) & 95 (88.7--98.4) & 93 (86.1--97.1) & 88 (80.0--93.6) \\
Multi-hop & 95 (88.7--98.4) & 62 (51.7--71.5) & 73 (63.2--81.4) & 36 (26.6--46.2) \\
Both hops & 97.5 (94.3--99.2) & 78.5 (72.2--84.0) & 83.0 (77.1--87.9) & 62.0 (54.9--68.8) \\
ABSTAIN (SH / MH) & 0 / 0 & 2 / 7 & 5 / 4 & 4 / 10 \\
\bottomrule
\end{tabular}
\end{table}

\begin{table}[htbp]
\centering
\footnotesize
\caption{Residual label by question set (8 sets = 4 fact lists $\times$ \{multi-hop, single-hop\}); $n=100$ questions per set. Counts from the row-level audit file.}
\label{tab:crosstab}
\begin{tabular}{@{}lrrrr@{}}
\toprule
Question set & Failures & Last-write-reachable miss & Ambiguous gold & Parser scope \\
\midrule
Multi-hop 6K & 5 & 1 & 4 & 0 \\
Multi-hop 32K & 38 & 22 & 13 & 3 \\
Multi-hop 64K & 27 & 15 & 9 & 3 \\
Multi-hop 262K & 64 & 29 & 34 & 1 \\
Single-hop 6K & 0 & 0 & 0 & 0 \\
Single-hop 32K & 5 & 2 & 2 & 1 \\
Single-hop 64K & 7 & 3 & 2 & 2 \\
Single-hop 262K & 12 & 6 & 3 & 3 \\
\midrule
All & 158 & 78 & 67 & 13 \\
\bottomrule
\end{tabular}
\end{table}

\section{Published comparison: all cited rows}\label{app:published}

All values are transcribed from \citet{hu2025memoryagentbench}, version v1 (arXiv, 7 Jul 2025), and are not recomputed; they are percentages under the benchmark's accuracy / substring exact match (SubEM) mapping.

\begin{table}[htbp]
\centering
\footnotesize
\setlength{\tabcolsep}{4pt}
\caption{Tier-matched comparison used in \S\ref{sec:published}: cited published Conflict Resolution scores beside the resolver, by tier; transcribed from \citet{hu2025memoryagentbench} v1 (last row: v4 camera-ready), not recomputed. Main table = the benchmark's v1 Table~2 / App.\ Table~7 (262K tier); v1 App.\ Table~11 = per-tier long-context rows; v1 Table~3 = its reasoning-model run on short lists.}
\label{tab:pub}
\begin{tabular}{@{}L{1.83in}L{1.1in}L{0.8in}L{1.42in}@{}}
\toprule
System & Tier & Single-hop (\%) & Multi-hop (\%) \\
\midrule
Frozen last-write resolver (this work) & 6K (dev) / 32K / 64K / 262K & 100 / 95 / 93 / 88 & 95 / 62 / 73 / 36 \\
GPT-4o, long-context (the benchmark's v1 Tables~3, 11, 2) & 6K / 32K / 64K / 262K & 92 / 88 / 85 / 60 & 28 / 10 / 13 / 5 \\
o4-mini, reasoning model (v1 Table~3) & 6K / 32K & 100 / 61 & 80 / 14 \\
Best RAG or memory system, v1 main table & 262K & 56 (BM25) & 7 (Contriever) \\
GPT-5-mini, long-context (v4) & 262K & 78 & 28 \\
\bottomrule
\end{tabular}
\end{table}

The benchmark's own runs cap FactConsolidation output at 10 tokens (v1 Table~14) and use a chunk size of 512 (v1 Table~15).
The ICLR 2026 camera-ready (arXiv v4, 28 Jun 2026) renames the competency Selective Forgetting, relabels the metric Accuracy (still substring exact match), revises BM25 single-hop from 56 to 48, drops NV-Embed-v2, and adds GPT-5-mini 78 / 28, Qwen3-Embedding-4B 29 / 3, MemoRAG 21 / 7, Zep 7 / 3, MIRIX 14 / 2 and MIRIX (4.1-mini) 20 / 3 at 262K; every other number cited here is unchanged except the per-tier correction noted next.
In v1 App.\ Table~11 the Claude-3.7-Sonnet multi-hop entry at 262K is 0 where v1 Table~2 / App.\ Table~7 give 2 for the same cell; the v4 camera-ready corrects the per-tier table to 2.

\begin{table}[htbp]
\centering
\footnotesize
\caption{All 17 systems of the benchmark's v1 main results table (v1 Table~2 / App.\ Table~7; 262K tier). Categories as in the source; all non-long-context rows use a GPT-4o-mini backbone.}
\label{tab:pub17}
\begin{tabular}{@{}L{2.35in}L{1.55in}rr@{}}
\toprule
System & Category & Single-hop & Multi-hop \\
\midrule
GPT-4o & long-context & 60 & 5 \\
GPT-4o-mini & long-context & 45 & 5 \\
GPT-4.1-mini & long-context & 36 & 5 \\
Gemini-2.0-Flash & long-context & 30 & 3 \\
Claude-3.7-Sonnet & long-context & 43 & 2 \\
BM25 & simple RAG & 56 & 3 \\
Contriever & embedding RAG & 18 & 7 \\
Text-Embed-3-Small & embedding RAG & 28 & 3 \\
Text-Embed-3-Large & embedding RAG & 28 & 4 \\
NV-Embed-v2 & embedding RAG & 55 & 6 \\
RAPTOR & structure-augmented RAG & 14 & 1 \\
GraphRAG & structure-augmented RAG & 14 & 2 \\
HippoRAG-v2 & structure-augmented RAG & 54 & 5 \\
Mem0 & structure-augmented RAG & 18 & 2 \\
Cognee & structure-augmented RAG & 28 & 3 \\
Self-RAG & agentic memory & 19 & 3 \\
MemGPT & agentic memory & 28 & 3 \\
\midrule
Frozen last-write resolver (this work, 262K list only) & deterministic, no learned component & 88 & 36 \\
\bottomrule
\end{tabular}
\end{table}

Pooled over all four lists the resolver scores 94.0 single-hop / 66.5 multi-hop (92.0 / 57.0 on the three held-out lists); only its 262K row is tier-matched to the main table.
At 6K, the source's own reasoning-model run (o4-mini, v1 Table~3) reaches 100 / 80, within 15 pp of the resolver's 95 multi-hop on the development list.
The camera-ready's App.~K.1 states the reading that the item-level analysis in this paper qualifies: ``We clarify that while prompt deltas can shift outcomes, our benchmark applies a unified and standardized prompt template across all evaluated agents (Long-context, RAG, and Agentic). This ensures that the performance gaps observed (e.g., the failure of RAG agents in multi-hop Selective Forgetting) are attributed to the limitations of their memory mechanisms rather than prompt inconsistency.''
The camera-ready's App.~G, item 4, states the same reading from the 6K runs: ``As demonstrated in our ablation study, long-context agents with strong reasoning capabilities achieve near-perfect performance on short-context (6K tokens) versions of the dataset. This result directly confirms that performance degradation on long-context, full-length task inputs stems not from flaws in the task definition itself, but from the fundamental limitations of current memory agents in long-range reasoning---specifically, their inability to accurately identify and discard outdated information across extended interaction histories.''
On this split a gap can also reflect items whose gold the stated rule does not reach (\S\ref{sec:residual}) and the failure at 262K of both long-context models on items the rule solves (\S\ref{sec:lc}); the item-level analysis qualifies both readings.

\begin{table}[htbp]
\centering
\footnotesize
\caption{Per-tier published rows. Top: v1 Table~3 (``Performances of reasoning models on the dataset FactConsolidation''). Bottom: v1 App.\ Table~11 (``Performance comparison on different context length''), FactConsolidation columns only.}
\label{tab:pubtier}
\begin{tabular}{@{}lrrrr@{}}
\toprule
 & \multicolumn{2}{c}{Single-hop} & \multicolumn{2}{c}{Multi-hop} \\
v1 Table~3 & 6K & 32K & 6K & 32K \\
\midrule
GPT-4o & 92 & 88 & 28 & 10 \\
o4-mini & 100 & 61 & 80 & 14 \\
\bottomrule
\end{tabular}

\medskip

\begin{tabular}{@{}lrrrrrr@{}}
\toprule
 & \multicolumn{3}{c}{Single-hop} & \multicolumn{3}{c}{Multi-hop} \\
v1 App.\ Table~11 & 32K & 64K & 262K & 32K & 64K & 262K \\
\midrule
GPT-4o & 88 & 85 & 60 & 10 & 13 & 5 \\
GPT-4o-mini & 63 & 58 & 45 & 10 & 5 & 5 \\
GPT-4.1-mini & 82 & 72 & 36 & 7 & 9 & 5 \\
Gemini-2.0-Flash & 49 & 62 & 30 & 7 & 9 & 3 \\
Claude-3.7-Sonnet & 46 & 45 & 43 & 2 & 2 & 0 \\
Mem0 & 22 & 8 & 18 & 3 & 2 & 2 \\
Cognee & 39 & 31 & 28 & 4 & 5 & 3 \\
\midrule
Frozen last-write resolver (this work) & 95 & 93 & 88 & 62 & 73 & 36 \\
\bottomrule
\end{tabular}
\end{table}

\section{Freeze and reproducibility record}\label{app:freeze}

\begin{itemize}
\item Dataset: the released Conflict Resolution parquet, SHA-256 \texttt{24d5c3f0}\ldots\ (1,491,588 bytes); 8 rows, 4 distinct context strings, 800 questions.
\item Development rows 0 and 4 (the 6K fact list with its multi-hop and single-hop question sets) declared before any content was inspected; no question, answer or statement text from the other six rows was inspected before the freeze.
\item Frozen parser/resolver source: SHA-256 \texttt{cc4ea4e1}\ldots; freeze time \mbox{2026-09-02T00:25:35Z}; the all-row evaluation refuses to run unless the source hash matches the freeze record; no edits after the freeze. Development-set diagnostic before the freeze: 195/200 correct, 0 ABSTAIN, 910/910 statements parsed.
\item Parse check: the pipeline's script recorded a 30-statement sample (seed 20260901) as faithful without a per-row check (all 30 rows marked correct by the script); a later re-application of the same patterns to the same sample, drawn from the parser's own output, reproduces the tuples by construction and is a consistency check that cannot fail; no independent semantic audit of parses exists. The same seed is used for the clustered bootstrap.
\item Resolver execution and metric scoring (segmentation, parsing, graph construction, resolution, scoring, workbench construction, classification script) are local deterministic Python and make no LLM API calls; the 13 parser-scope designations consumed by the classification script were made in the LLM-assisted batch review (App.~\ref{app:rule}), and the long-context experiment of \S\ref{sec:lc} retains one scored output per question (retries: App.~\ref{app:lc}).
\item Scoring: the official repository's \texttt{substring\_exact\_match} at commit \texttt{fe1735de}, re-implemented verbatim---lowercase, strip ASCII punctuation, drop the articles \emph{a}/\emph{an}/\emph{the}, collapse whitespace, then test whether any normalized gold string is a substring of the normalized prediction; strict case-sensitive equality reported alongside (identical counts in every group).
\item Erratum: the frozen record \texttt{parser-freeze.md} says 32 patterns; the frozen source has 35 patterns mapping to 34 relation keys (App.~\ref{app:parser}).
\item Novelty check against the source paper (v1, 23 pages, and the v4 camera-ready) and the official repository at commit \texttt{fe1735de}: neither version contains a deterministic zero-learning baseline on this split or a row-level residual taxonomy; the nearest construction elsewhere, StateMemBench \citep{fan2026statemem}, generates its own scenarios rather than auditing an existing benchmark's items; LongMemEval's knowledge-update questions \citep{wu2025longmemeval} are the natural next target for the instrument.
\item Available from the authors on request: the frozen source, the per-question results, the parsed and unparsed statement files, the 158-row workbench and labeled audit file, and the classification script.
\end{itemize}

\section{LLM-use disclosure}\label{app:llm}

Language models enter this work in two roles. As instruments, in three labeling steps disclosed in App.~\ref{app:rule}---the batch review that designated the 13 parser-scope rows (GPT-5.6) and the two blind re-labelings (Claude Opus 4.8, re-adjudication A; Gemini 3.1 Pro (preview), re-adjudication B)---and in the row-by-row check of the 67 ambiguous-gold rows (Claude Opus 4.8 and GPT-5.6, independently). As subjects: the long-context models of \S\ref{sec:lc}, Gemini 2.5 Pro and Gemini 2.5 Flash, and the retrieval agent's GPT-4o-mini backbone.
LLM assistance was also used in the engineering of the resolver and analysis code and in manuscript preparation, under the authors' direction and review; the authors take full responsibility for all content, claims, and measurements.
Parsing, resolution and scoring were produced by scripted pipelines with no LLM call.
The pipeline's audit script recorded a 30-statement parse sample as faithful without a per-row check, and a later re-application of the same patterns to the same sample is a consistency check that cannot fail; no independent semantic audit of parses exists (App.~\ref{app:freeze}).
The 158 residual-failure labels were produced by the classification script over a row-level evidence workbench: the 13 designated rows are an explicit list in the script; every other row was labeled by the script from the workbench's last-write reachability flag (gold reachable in the frozen last-write graph $\to$ last-write-reachable miss; not reachable $\to$ ambiguous gold); the script's explicit override lists for the categories genuine beyond last write, normalization artifact and unclassifiable were left empty.
The ambiguous-gold labels are corroborated by the programmatic reachability check and the row-by-row check in App.~\ref{app:rule}.
The re-derivation script that reproduces every reported number from the frozen artifacts, its output, and the reference-verification record are available from the authors on request.

\section{Long-context and retrieval-agent experiments: protocol and tables}\label{app:lc}

\textbf{Design (pre-registered before any call; amendments and deviations listed below).}
All 800 questions, one retained output per question (retry protocol below), in the benchmark's long-context setting reproduced from the official repository at commit \texttt{fe1735de} (\texttt{utils/templates.py}, \texttt{BASE\_TEMPLATES['factconsolidation']}; the long-context agent in \texttt{agent.py}): the system message ``You are a helpful assistant that can read the context and memorize it for future retrieval.''; the fact list split at line boundaries into chunks of about 16,000 characters ($\approx$4,096 tokens, the repository's configured chunk size at that commit; the paper's v1 Table~15 lists 512), each wrapped in the repository's memorize template (``Dialogue between User and Assistant \ldots\ <User> The following context is the facts I have learned: \ldots\ <Assistant> I have learned the facts and I will answer the question you ask.'') with a fixed time stamp; then the repository's query template, whose rule sentence reads ``Each fact in the knowledge pool is provided with a serial number at the beginning, and the newer fact has larger serial number. You need to solve the conflicts of facts in the knowledge pool by finding the newest fact with larger serial number.'', followed by the question and ``Answer:''.
Fact list first, question last; no tools.
The paper's \S\ref{sec:intro} quotes the benchmark paper's Figure~5 wording, which ends at ``finding the newest fact''; the repository template used here adds ``with larger serial number''.

\textbf{Model selection.}
Fixed before any result by a cost rule: candidates in the order gpt-5.5, claude-sonnet-5, gemini-2.5-pro, gemini-2.5-flash; a two-call smoke test per candidate (one 6K item, one 262K item); the primary model is the first in that order whose projected 800-item cost is at or below a pre-set budget; a cheaper secondary model (gemini-2.5-flash) is run as a robustness check if it fits a smaller pre-set budget.
The rule selected Gemini 2.5 Pro as primary and Gemini 2.5 Flash as secondary; both were run under the identical protocol.

\textbf{Sampling and output budget.}
Temperature is the provider default (a fixed value of 0.0 was rejected by two of the four candidate models), so outputs are not deterministic; one run per item per model, no best-of-$n$.
Each item was requested with up to three attempts at a 4,000-token output budget; empty outputs (hidden reasoning exhausting the budget) and transport/rate-limit errors were retried, non-empty outputs were never discarded; items still unresolved---11 for Pro (6 empty, 5 transport) and 17 for Flash (all transport)---were re-requested under the same loop with a 16,000-token budget, under an amendment declared before the retry; two smoke-test items per candidate model preceded the run and are not scored. Single runs vary run to run even at temperature 0 \citep{bjarnason2026randomness,atil2024nondeterminism}, and across inference configurations for numerical reasons \citep{yuan2025numerical}; the repeat-run analysis below measures the former for the retrieval agent. The long-context runs are the full-context baseline that memory-system evaluations are advised to report \citep{jiang2026anatomy}; this paper audits one update workload and harness and compares no memory architectures \citep{weready2026memory}. 
\paragraph{Repeat runs.} Under pre-registration amendment R-A1 the BM25 agent was run three times under the identical protocol (the retained first run plus two repeats; temperature 0.7). Long-context amendments A3/A4 pre-registered a repeat of Gemini 2.5 Flash; that run was stopped at 368 of 800 items after provider errors and rate limiting cut its throughput (amendment A5, decided before any of its results were analyzed), so the two Gemini models are reported as single runs. Table~\ref{tab:repeat} gives per-class accuracy per run for the agent; the range across runs is the run-to-run variation a single run hides.
\begin{table}[htbp]\centering\footnotesize\setlength{\tabcolsep}{4pt}\caption{Repeat runs: accuracy (\%) per item class and run; last two columns are the per-item agreement between runs ($\kappa$; the three pairwise values round to the same number) and the number of items whose outcome differs between a pair of runs (range).}\label{tab:repeat}
\begin{tabular}{llrrrrr}\toprule System & class & run 1 & run 2 & run 3 & $\kappa$ (pairs) & differing items \\ \midrule
BM25 + GPT-4o-mini & resolver-correct & 41.6 & 40.3 & 41.7 & 0.90 & 35--37 \\
 & last-write-reachable miss & 12.8 & 10.3 & 11.5 &  &  \\
 & ambiguous gold & 6.0 & 9.0 & 9.0 &  &  \\
 & parser scope & 30.8 & 23.1 & 7.7 &  &  \\
 & all items & 35.6 & 34.5 & 35.5 &  &  \\
\bottomrule\end{tabular}\end{table}

The run files contain 811 (Pro) and 817 (Flash) records for 800 items; the long-context pre-registration's amendment A2 described the 11 Pro retries as empty outputs, the run file shows 6 empty and 5 transport failures, and a dated note in the pre-registration records the correction.
Both models returned a visible answer for 800 of 800 items (every finish reason ``stop''); provider-reported completion tokens, including hidden reasoning, average 554 (Pro) and 439 (Flash), while the visible outputs average 1.8 and 2.3 words.
The retrieved template and agent source files are kept alongside the prompt file in the analysis record.

\textbf{Scoring.}
The official \texttt{substring\_exact\_match} applied (a) to the full visible output and (b) to the first line truncated to 10 words (the benchmark caps generation at 10 tokens).
For Pro, (a) and (b) give identical counts on every group; for Flash they differ on one item (593 vs.\ 592 correct overall).
The pre-registration named scoring (b) as the headline; the paper reports (a) because the two differ on one Flash item and (a) is the alignment quantity; both are in the scored files.
Every deviation from the pre-registration: the repository's system message was used (the pre-registration said none); the output budget was raised from 200 to 4,000 tokens with a single 16,000-token retry for still-empty or errored items; and scoring (a) rather than (b) is the headline; temperature is the provider default as pre-registered (the smoke test had sent a fixed 0.0, which two candidates rejected).
Also recorded, under the same normalization: whether the output contains the resolver's last-write prediction for the item.

\begin{table}[htbp]
\centering
\footnotesize
\setlength{\tabcolsep}{4pt}
\caption{Accuracy (\%) per question set and pooled under the official metric, with exact Clopper--Pearson 95\% CIs, for both models.}
\label{tab:lcsets}
\begin{tabular}{@{}lrrlrl@{}}
\toprule
Group & $n$ & Pro correct & Pro accuracy (95\% CI) & Flash correct & Flash accuracy (95\% CI) \\
\midrule
Single-hop 6K & 100 & 99 & 99.0 (94.6--100.0) & 100 & 100.0 (96.4--100.0) \\
Single-hop 32K & 100 & 98 & 98.0 (93.0--99.8) & 96 & 96.0 (90.1--98.9) \\
Single-hop 64K & 100 & 98 & 98.0 (93.0--99.8) & 95 & 95.0 (88.7--98.4) \\
Single-hop 262K & 100 & 65 & 65.0 (54.8--74.3) & 77 & 77.0 (67.5--84.8) \\
Multi-hop 6K & 100 & 90 & 90.0 (82.4--95.1) & 83 & 83.0 (74.2--89.8) \\
Multi-hop 32K & 100 & 70 & 70.0 (60.0--78.8) & 61 & 61.0 (50.7--70.6) \\
Multi-hop 64K & 100 & 71 & 71.0 (61.1--79.6) & 56 & 56.0 (45.7--65.9) \\
Multi-hop 262K & 100 & 9 & 9.0 (4.2--16.4) & 25 & 25.0 (16.9--34.7) \\
\midrule
Single-hop, all four lists & 400 & 360 & 90.0 (86.6--92.8) & 368 & 92.0 (88.9--94.5) \\
Multi-hop, all four lists & 400 & 240 & 60.0 (55.0--64.8) & 225 & 56.3 (51.2--61.2) \\
Held-out (32K--262K), both hops & 600 & 411 & 68.5 (64.6--72.2) & 410 & 68.3 (64.4--72.0) \\
All & 800 & 600 & 75.0 (71.8--78.0) & 593 & 74.1 (70.9--77.1) \\
\bottomrule
\end{tabular}
\end{table}

\begin{table}[htbp]
\centering
\footnotesize
\caption{Model outcome (correct / wrong) by resolver item class, overall and per fact list, for both models. No item is missing.}
\label{tab:lccross}
\begin{tabular}{@{}llrrrrr@{}}
\toprule
Model & Item class & All & 6K & 32K & 64K & 262K \\
\midrule
Pro & Resolver-correct (642) & 544 / 98 & 188 / 7 & 141 / 16 & 152 / 14 & 63 / 61 \\
 & Last-write-reachable miss (78) & 41 / 37 & 1 / 0 & 20 / 4 & 13 / 5 & 7 / 28 \\
 & Ambiguous gold (67) & 7 / 60 & 0 / 4 & 3 / 12 & 1 / 10 & 3 / 34 \\
 & Parser scope (13) & 8 / 5 & 0 / 0 & 4 / 0 & 3 / 2 & 1 / 3 \\
 & All (800) & 600 / 200 & 189 / 11 & 168 / 32 & 169 / 31 & 74 / 126 \\
\midrule
Flash & Resolver-correct (642) & 530 / 112 & 182 / 13 & 133 / 24 & 131 / 35 & 84 / 40 \\
 & Last-write-reachable miss (78) & 46 / 32 & 1 / 0 & 19 / 5 & 14 / 4 & 12 / 23 \\
 & Ambiguous gold (67) & 8 / 59 & 0 / 4 & 2 / 13 & 2 / 9 & 4 / 33 \\
 & Parser scope (13) & 9 / 4 & 0 / 0 & 3 / 1 & 4 / 1 & 2 / 2 \\
 & All (800) & 593 / 207 & 183 / 17 & 157 / 43 & 151 / 49 & 102 / 98 \\
\bottomrule
\end{tabular}
\end{table}

\textbf{Accuracy by class with intervals.}
Pro: resolver-correct items 84.7\% (81.7--87.4; per list 96.4 / 89.8 / 91.6 / 50.8 at 6K / 32K / 64K / 262K); last-write-reachable-miss items 52.6\% (40.9--64.0; 100 / 83.3 / 72.2 / 20.0 on 1 / 24 / 18 / 35 items); ambiguous-gold items 10.4\%; parser-scope items 61.5\% (31.6--86.1); shares of its 200 errors 49.0 / 18.5 / 30.0 / 2.5\% in that class order.
Flash: resolver-correct items 82.6\% (79.4--85.4; per list 93.3 / 84.7 / 78.9 / 67.7); last-write-reachable-miss items 59.0\% (47.3--70.0; 100 / 79.2 / 77.8 / 34.3); ambiguous-gold items 11.9\%; parser-scope items 69.2\% (38.6--90.9); shares of its 207 errors 54.1 / 15.5 / 28.5 / 1.9\% in that class order.

\textbf{The 67 operationally ambiguous items.}
Pro's output contains the resolver's last-write prediction in 17 of 67 (95\% CI 15.5--37.5\%), the released gold in 7 (4.3--20.3\%), both in 0, and neither in 43 (51.5--75.5\%); per list, outputs containing the last-write prediction: 6K 4 of 4, 32K 7 of 15, 64K 4 of 11, 262K 2 of 37.
Flash: last-write prediction in 16 (14.3--35.9\%), gold in 8 (5.3--22.2\%), both in 0, neither in 43 (51.5--75.5\%); per list 6K 2 of 4, 32K 6 of 15, 64K 3 of 11, 262K 5 of 37.
On the 78 last-write-reachable-miss items neither model's output ever contains the resolver's (wrong) prediction; on the 13 parser-scope items each does once.

\textbf{Two-model agreement.}
The two models' per-item outcomes agree on 657 of 800 items (82.1\%), Cohen's $\kappa = 0.53$.
Table~\ref{tab:lctwo} gives the cross-tab by item class: on the 67 ambiguous-gold items both are wrong on 53, on the 642 resolver-correct items both are right on 484.

\textbf{Retrieval agent (Experiment R).}
Our pre-registered approximate re-implementation of the benchmark's BM25 configuration (BM25 over 512-token chunks, top 10, GPT-4o-mini; the main table's Simple RAG + BM25 row, 56 / 3 at 262K in v1, 48 / 3 in v4) was run on all 800 questions under a protocol mirroring the repository at commit \texttt{fe1735de} as far as documented.
Ingestion: the fact list is split into consecutive 512-token windows (\texttt{tiktoken} \texttt{o200k\_base}; the paper's v1 Table~15 chunk size, the yaml says 4,096); the official chunker is not in the snapshot, so windows may split inside a fact line; each chunk is wrapped in the memorize template (fixed time stamp), and the wrapped text is what BM25 indexes (12 / 64 / 128 / 524 chunks per list at 6K / 32K / 64K / 262K).
Retrieval: \texttt{BM25Okapi} over lowercase whitespace tokens with the question text as query; the top 10 chunks in rank order, each prefixed ``Memory $i$:'' as in the repository's agent code (the top 10 cover about 83\% / 16\% / 8\% / 2\% of the four lists).
Generation: the repository's RAG query template and system message; temperature 0.7 (the benchmark's setting); maximum output 200 tokens; one retained output per question with the same retry loop as above (800 items, 800 returned, one retry used, mean completion length 2.8 tokens); two smoke-test items preceded the run and are not scored.
Four departures from the official code: lowercased BM25 tokens and the bare question as query (both pre-registered; the official retriever uses \texttt{text.split()} without lowercasing and a longer query string), a guessed chunker that may split statements (the official chunker is not in the snapshot), and a 200-token output budget; offline, the top-10 chunk set equals the closest official-code variant's on 134 of 800 items (ranked order on 3), with hit rates within about 2 points.
Scoring (a) as above; the agent scores 285/800 = 35.6\% (95\% CI 32.3--39.1); held-out 196/600 = 32.7\% (28.9--36.6); single-hop 251/400 = 62.8\% (57.8--67.5); multi-hop 34/400 = 8.5\% (6.0--11.7); per set (single-hop / multi-hop, correct of 100): 6K 70 / 19, 32K 65 / 7, 64K 60 / 4, 262K 56 / 4, with 95\% CIs 60.0--78.8 / 11.8--28.1, 54.8--74.3 / 2.9--13.9, 49.7--69.7 / 1.1--9.9, 45.7--65.9 / 1.1--9.9; its 262K row (56 / 4) sits beside the benchmark's published 56 / 3---a coarse fidelity check, not a replication.

\begin{table}[htbp]
\centering
\footnotesize
\caption{Retrieval agent (BM25 + GPT-4o-mini): outcome (correct / wrong) by resolver item class, overall and per fact list. No item is missing.}
\label{tab:ragcross}
\begin{tabular}{@{}lrrrrr@{}}
\toprule
Item class & All & 6K & 32K & 64K & 262K \\
\midrule
Resolver-correct (642) & 267 / 375 & 89 / 106 & 66 / 91 & 59 / 107 & 53 / 71 \\
Last-write-reachable miss (78) & 10 / 68 & 0 / 1 & 3 / 21 & 3 / 15 & 4 / 31 \\
Ambiguous gold (67) & 4 / 63 & 0 / 4 & 1 / 14 & 1 / 10 & 2 / 35 \\
Parser scope (13) & 4 / 9 & 0 / 0 & 2 / 2 & 1 / 4 & 1 / 3 \\
\midrule
All (800) & 285 / 515 & 89 / 111 & 72 / 128 & 64 / 136 & 60 / 140 \\
\bottomrule
\end{tabular}
\end{table}

Per class: resolver-correct items 41.6\% (37.7--45.5), last-write-reachable misses 12.8\%, parser-scope items 30.8\%, ambiguous-gold items 6.0\% (4/67); shares of its 515 errors 72.8 / 13.2 / 12.2 / 1.7\%.
On the 67 ambiguous-gold items its output contains the resolver's last-write prediction in 4, the released gold in 4, and neither in 59.
Per-item outcome agreement with the long-context models is low: $\kappa = 0.20$ with Pro and $0.23$ with Flash.
Retrieval diagnostic (no LLM call): for 496 of the 800 items every statement the resolver used lies in the agent's top-10 chunks; on those items the agent scores 262/496 = 52.8\% (48.3--57.3).
Hop stratification: of the 67 operationally ambiguous items 60 are multi-hop, of the 642 resolver-correct items 376 are single-hop; within multi-hop items, accuracy on resolver-correct (266) vs.\ operationally ambiguous (60) items is 198/266 (74.4\%) vs.\ 5/60 (8.3\%) for Pro, 176/266 (66.2\%) vs.\ 7/60 (11.7\%) for Flash and 31/266 (11.7\%) vs.\ 0/60 for the agent; at multi-hop 262K (36 resolver-correct, 34 ambiguous items), 4/36 vs.\ 1/34, 13/36 vs.\ 4/34 and 4/36 vs.\ 0/34.

\begin{table}[htbp]
\centering
\footnotesize
\caption{Per-item outcome cross-tab of the two models (Pro outcome $\times$ Flash outcome) by resolver item class.}
\label{tab:lctwo}
\begin{tabular}{@{}lrrrr@{}}
\toprule
Item class & Both correct & Pro only & Flash only & Both wrong \\
\midrule
Resolver-correct (642) & 484 & 60 & 46 & 52 \\
Last-write-reachable miss (78) & 33 & 8 & 13 & 24 \\
Ambiguous gold (67) & 1 & 6 & 7 & 53 \\
Parser scope (13) & 7 & 1 & 2 & 3 \\
\midrule
All (800) & 525 & 75 & 68 & 132 \\
\bottomrule
\end{tabular}
\end{table}

\end{document}